\documentclass[runningheads]{llncs}
\usepackage[T1]{fontenc}
\usepackage{graphicx}
\usepackage{url}
\usepackage{amssymb}
\usepackage{amsmath}
\usepackage{multirow}
\usepackage{amsfonts}
\usepackage{xcolor}
\usepackage{textcomp}
\usepackage{manyfoot}
\usepackage{booktabs}
\usepackage{subfig}
\usepackage{algpseudocode}
\usepackage{cleveref}
\usepackage{lmodern}
\usepackage{algorithm}
\usepackage{algorithmicx}
\usepackage{arydshln}
\usepackage{stfloats}
\usepackage{verbatim}

\begin{document}
\title{MsEdF: A Multi-stream Encoder-decoder Framework for Remote Sensing Image Captioning}
%
%
\author{Swadhin Das\inst{1}\orcidID{0009-0004-1247-3275} \and
Raksha Sharma\inst{1}\orcidID{0000-0003-2905-0194}
}
\authorrunning{Das et al.}
%
\institute{
Indian Institute of Techology Roorkee\\
\email{\{s\_das,raksha.sharma\}@cs.iitr.ac.in}}
\maketitle              
\begin{abstract}
Remote sensing images contain complex spatial patterns and semantic structures, which makes the captioning model difficult to accurately describe.  Encoder-decoder architectures have become the widely used approach for RSIC by translating visual content into descriptive text.  However, many existing methods rely on a single-stream architecture, which weakens the model to accurately describe the image. Such single-stream architectures typically struggle to extract diverse spatial features or capture complex semantic relationships, limiting their effectiveness in scenes with high intraclass similarity or contextual ambiguity. In this work, we propose a novel Multi-stream Encoder-decoder Framework (MsEdF) which improves the performance of RSIC by optimizing both the spatial representation and language generation of encoder-decoder architecture. The encoder fuses information from two complementary image encoders, thereby promoting feature diversity through the integration of multiscale and structurally distinct cues. To improve the capture of context-aware descriptions, we refine the input sequence's semantic modeling on the decoder side using a stacked GRU architecture with an element-wise aggregation scheme.  Experiments on three benchmark RSIC datasets show that MsEdF outperforms several baseline models. 

\end{abstract}
\section{Introduction}
\label{Introduction}

Remote sensing image captioning (RSIC) refers to the task of automatically generating descriptive sentences from high‑resolution remote sensing images. A widely adopted approach for generating accurate captions from remote sensing images is the encoder–decoder framework. Here a convolutional neural network (CNN) extracts semantic features and a sequential model, typically an RNN or LSTM, generates textual descriptions \cite{qu2016deep,lu2017exploring}. Most existing approaches for RSIC still rely on conventional single-stream encoder and decoder architectures, which often struggle to model the complex spatial composition and semantic ambiguity in remote sensing images \cite{hoxha2021novel,hoxha2020new}. These models typically adopt a single CNN encoder, such as ResNet \cite{das2024textgcn}, InceptionNet \cite{hoxha2020new}, or VGGNet \cite{lu2017exploring}, without exploring how encoder diversity impacts captioning performance \cite{das2024unveiling}. However, single encoders are limited in capturing both global context and fine-grained object-level details due to constrained receptive fields and uniform feature hierarchies \cite{das2024textgcn}, which reduces the expressiveness of extracted visual features.

Decoder components exhibit similar bottlenecks. Many models use shallow decoders based on single-layer LSTM \cite{das2024unveiling}, GRU \cite{hoxha2020new}, or SVM \cite{hoxha2021novel}, which often overlook hierarchical structure and intermediate semantics, leading to repetitive or oversimplified captions. Deeper decoders have been introduced in related contexts, but their exploration has not been as extensive for RSIC~\cite{alghamdi2024predicting}. In addition, previous stacking architectures employ the final output layer alone~\cite{li2025feature}, which ignores intermediate signals with complementary semantic information. The opportunity to use adaptive fusion schemes, which can dynamically aggregate multiple-layer decoder output, has not been thoroughly studied.

To address these drawbacks, we design a multi-stream RSIC architecture~\cite{wu1509fusing}. On the encoder side, two different CNNs are fused to exploit global semantics and local fine-scale spatial information. On the decoder side, we propose a stacked GRU sequence generator that makes use of the output of the intermediate layers with a locally weighted fusion mechanism. This enables the decoder to integrate multilevel linguistic cues and generate more informative and coherent descriptions. Experiments on several benchmark datasets demonstrate that the proposed approach is competitive and consistent with an improvement over single-stream baselines.

The major contributions of this work are as follows.  
\begin{itemize}  
    \item A fusion-based encoding strategy is proposed to improve image representation and feature diversity by integrating distinct features from two complementary CNN encoders.  
    \item A novel multi-layer stacked GRU decoder with weighted averaging is proposed to improve semantic modeling of the input sequence.
\end{itemize}

The remainder of the paper is as follows. \Cref{sec:rel_work} reviews the literature based on our pre-work. \Cref{sec:prop_method} describes our proposed method. \Cref{sec:exp_setup} describes the experimental setup. \Cref{sec:results} presents the results. Finally, \Cref{sec:conclusion} summarizes our main findings and provides directions for future work.
\section{Related Work}
\label{sec:rel_work}
Remote Sensing Image Captioning (RSIC) has seen significant advances in recent years, using deep learning techniques to generate textual descriptions of remote sensing images. Although natural image capture (NIC) has been extensively studied \cite{you2016image}, RSIC remains a more specialized field with unique challenges due to its domain-specific characteristics \cite{hoxha2020new}. The most widely adoptd framework in this domain is the encoder-decoder architecture \cite{qu2016deep}, which forms the foundation of many state-of-the-art models.

Encoder-decoder architectures are commonly used in RSIC. Qu et al., [2016] \cite{qu2016deep} introduced a deep multimodal neural network for generating descriptions from remote sensing images. Lu et al., [2017] \cite{lu2017exploring} performed a comparative analysis on existing models using datasets such as SYDNEY \cite{qu2016deep} and UCM \cite{qu2016deep}, leading to the development of the RSICD dataset to address training data limitations. Das et al., [2024] \cite{das2024textgcn} introduced an encoder-decoder framework with a text graph convolutional network and multilayer LSTM for automated remote sensing image captioning, significantly improving caption generation by enhancing semantic understanding and using a comparison-based beam search. Liu et al., [2025] \cite{liu2025semantic} proposed a semantic-spatial feature fusion with dynamic graph refinement (SFDR) method for remote sensing image captioning, integrating multi-level feature representation and graph attention to improve the accuracy of generated descriptions. These methods significantly improved captioning accuracy, but still struggled with long-term dependencies and contextual consistency.

To overcome these limitations, attention mechanisms were introduced, significantly improving feature extraction and sequence modeling capabilities \cite{xu2015show}. Sattar et al., [2024] \cite{sattar2024remote} introduced the enhanced focal mechanism (EF) for image description generation, refining the focal point mechanism with multi-head attention and LSTM to improve correlation between image objects and improve subtitle quality. Wang et al., [2024] \cite{wang2024remote} presented a remote sensing image captioning model with sequential attention and flexible word correlation (SA-FWC), using VGG16 for feature fusion and LSTM for improved feature representation, achieving superior performance with minimal training sample pairs. Li et al. [2025] \cite{li2025feature} proposed a feature refinement and rethinking of the attention framework for remote sensing image captioning, incorporating a refinement gate and rethinking LSTM to improve region focus and model discriminative contextual representations.

Despite these improvements, existing models still suffer from inefficiencies in encoder utilization and suboptimal feature aggregation in the decoder. Many studies focus on refining the decoder \cite{hoxha2021novel,hoxha2020new}, but optimal feature extraction remains an open challenge. To address these issues, our approach leverages a fusion-based encoding strategy that integrates multiple CNN architectures for enhanced feature representation. Additionally, we incorporate locally weighted stacking of decoders to retain hierarchical linguistic patterns effectively.

\section{Proposed Method}
\label{sec:prop_method}
\begin{figure}[!ht]
    \centering
    \includegraphics[height=160px, width=\linewidth]{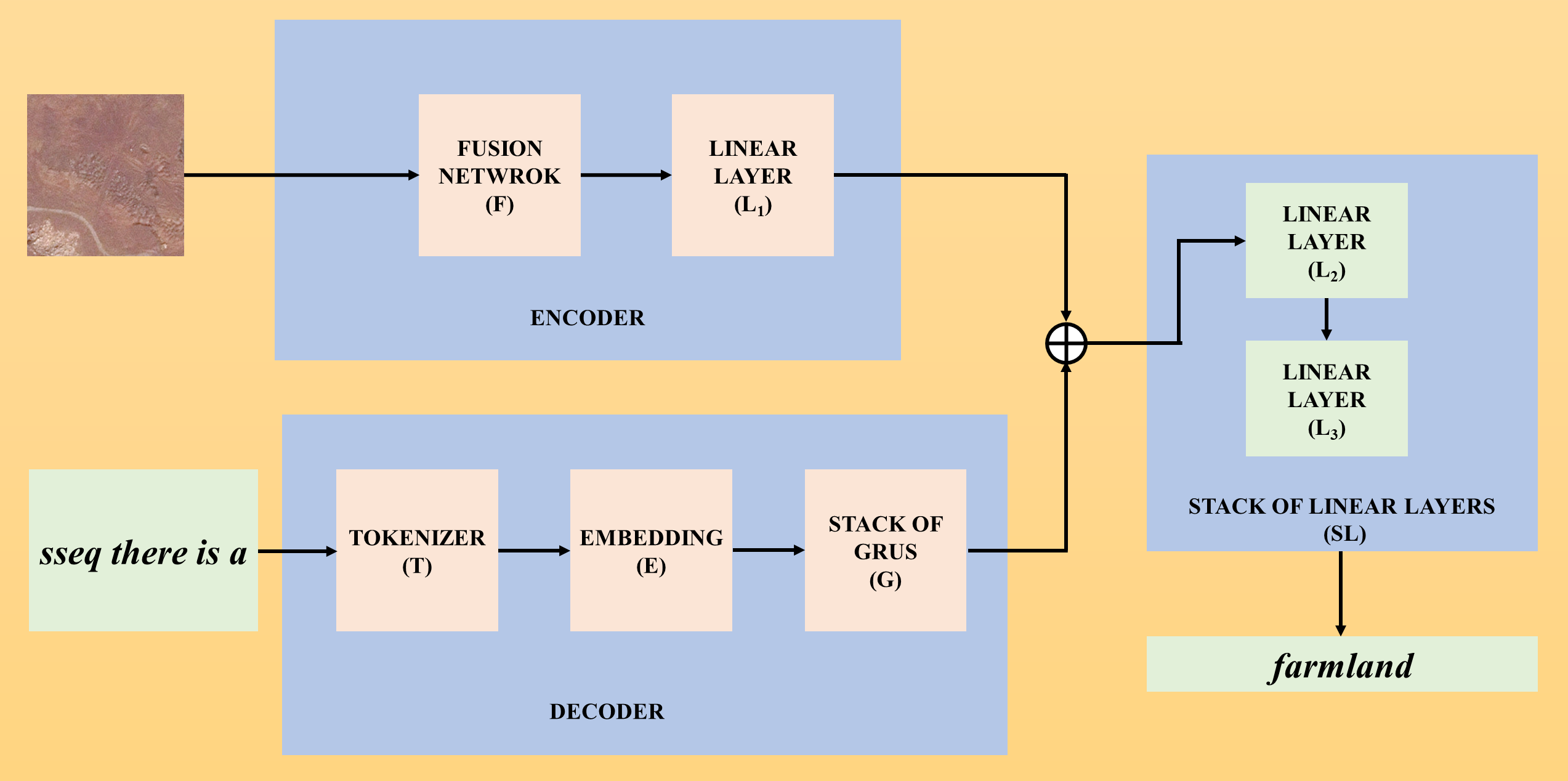}
    \caption{Architecture of the Proposed Model}
    \label{architecture}
\end{figure}
We follow an encoder-decoder architecture, where the encoder ($F$) encompasses two separate CNNs, whose outputs are concatenated and processed by a linear transformation layer ($L_1$). In the decoder, an input sequence in tokens is stored in an embedding layer ($E$) followed by series of GRU cells ($G$) to capture temporal relationships. The outputs of $L_1$ and $G$ are concatenated together, which is then used as input to two more linear layers $L_2$, and another layer $L_3$ serving as a softmax to generate word probabilities. The overall architecture of our work is shown in~\Cref{architecture}. 

\subsection{Heterogeneous Encoding Strategy for Feature Diversity}
\begin{figure}[!ht]
    \centering
    \includegraphics[height=135px, width=\linewidth]{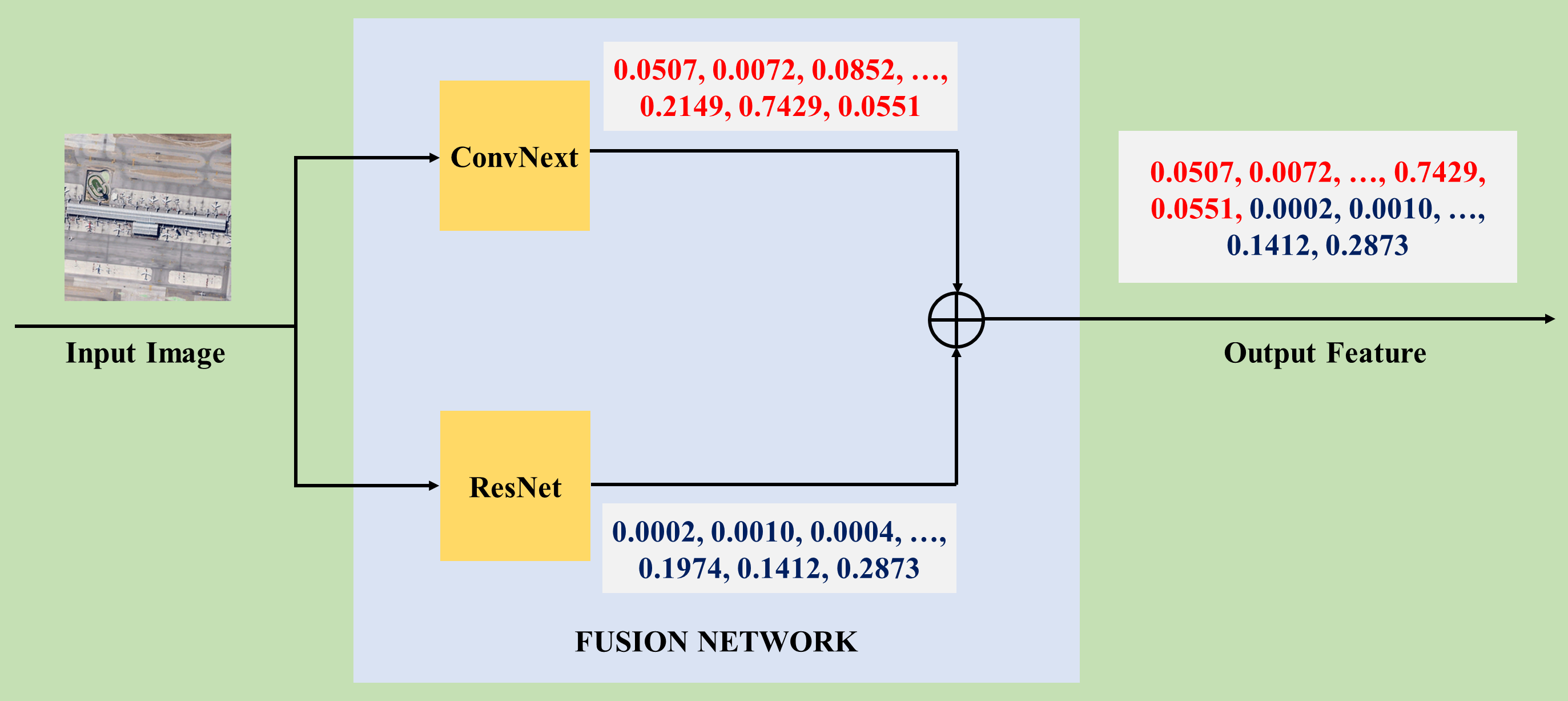}
    \caption{Illustrative Example of Encoder Fusion in the Proposed Model}
    \label{fusion}
\end{figure}
The proposed method integrates features extracted from two widely used CNN architectures. An illustrative example of the encoder fusion strategy is presented in~\Cref{fusion}. This study focuses on four prominent CNN architectures: ConvNext \cite{liu2022convnet}, ResNet \cite{he2016deep}, ResNext \cite{xie2017aggregated}, and DenseNet \cite{huang2017densely}. These models have demonstrated strong feature extraction capabilities in RSIC \cite{das2024unveiling,das2025good}.
\begin{table}[!ht]
\centering
\caption{Performance Comparison of Individual and Fusion-Based CNN Encoders on the SYDNEY Dataset}
\label{SYDNEY_CNN}
\resizebox{\linewidth}{!}{
    {\small
    \begin{tabular}{|c|c|c|c|c|c|c|c|}
    \hline
    CNN & BLEU-1 & BLEU-2 & BLEU-3 & BLEU-4 & METEOR & ROUGE-L & CIDEr \\
    \hline
    ConvNext & 0.7531 & 0.6645 & 0.5982 & 0.5376 & 0.3750 & 0.6951 & 2.3954 \\
    ResNet & 0.7363 & 0.6502 & 0.5785 & 0.5138 & 0.3621 & 0.6809 & 2.2526 \\
    ResNext & 0.7415 & 0.6556 & 0.5845 & 0.5209 & 0.3625 & 0.6706 & 2.2912 \\
    DenseNet & 0.7214 & 0.6331 & 0.5621 & 0.4935 & 0.3419 & 0.6314 & 2.1444 \\
    ConvNext-ResNet & \textbf{0.7679} & \textbf{0.6814} & \textbf{0.6084} & \textbf{0.5458} & 0.3801 & \textbf{0.7263} & \textbf{2.4515} \\
    ConvNext-ResNext & 0.7582 & 0.6727 & 0.6039 & 0.5438 & \textbf{0.3868} & 0.7164 & 2.3882 \\
    ConvNext-DenseNet & 0.7419 & 0.6571 & 0.5823 & 0.5264 & 0.3822 & 0.6911 & 2.3755 \\
    ResNet-ResNext & 0.7517 & 0.6664 & 0.5924 & 0.5243 & 0.3789 & 0.6963 & 2.3515 \\
    ResNet-DenseNet & 0.7246 & 0.6413 & 0.5722 & 0.5101 & 0.3644 & 0.6798 & 2.2488 \\
    ResNext-DenseNet & 0.7322 & 0.6514 & 0.5814 & 0.5299 & 0.3714 & 0.6833 & 2.2655 \\
    \hline
\end{tabular}}}
\end{table}
\begin{table}[!ht]
\centering
\caption{Performance Comparison of Individual and Fusion-Based CNN Encoders on the UCM Dataset}
\label{UCM_CNN}
\resizebox{\linewidth}{!}{
    {\small
    \begin{tabular}{|c|c|c|c|c|c|c|c|}
    \hline
    CNN & BLEU-1 & BLEU-2 & BLEU-3 & BLEU-4 & METEOR & ROUGE-L & CIDEr \\
    \hline
    ConvNext & 0.8369 & 0.7804 & 0.7312 & 0.6853 & 0.4689 & 0.8037 & 3.3990 \\
    ResNet & 0.8297 & 0.7705 & 0.7178 & 0.6683 & 0.4506 & 0.7903 & 3.3329 \\
    ResNext & 0.8170 & 0.7607 & 0.7118 & 0.6658 & 0.4583 & 0.7736 & 3.3165 \\
    DenseNet & 0.8222 & 0.7648 & 0.7211 & 0.6714 & 0.4674 & 0.7833 & 3.3419 \\
    ConvNext-ResNet & \textbf{0.8457} & \textbf{0.7919} & \textbf{0.7533} & \textbf{0.7166} & \textbf{0.4833} & \textbf{0.8163} & 3.4255 \\
    ConvNext-ResNext & 0.8406 & 0.7796 & 0.7289 & 0.6813 & 0.4656 & 0.7996 & 3.4132 \\
    ConvNext-DenseNet & 0.8416 & 0.7834 & 0.7319 & 0.6877 & 0.4798 & 0.8101 & \textbf{3.4522} \\
    ResNet-ResNext & 0.8334 & 0.7739 & 0.7236 & 0.6723 & 0.4580 & 0.7944 & 3.4108 \\
    ResNet-DenseNet & 0.8311 & 0.7825 & 0.7230 & 0.6711 & 0.4596 & 0.7998 & 3.3712 \\
    ResNext-DenseNet & 0.8367 & 0.7898 & 0.7292 & 0.6736 & 0.4674 & 0.8112 & 3.3919 \\
    \hline
\end{tabular}}}
\end{table}
\begin{table}[!ht]
\centering
\caption{Performance Comparison of Individual and Fusion-Based CNN Encoders on the RSICD Dataset}
\label{RSICD_CNN}
\resizebox{\linewidth}{!}{
    {\small
    \begin{tabular}{|c|c|c|c|c|c|c|c|}
    \hline
    CNN & BLEU-1 & BLEU-2 & BLEU-3 & BLEU-4 & METEOR & ROUGE-L & CIDEr \\
    \hline
    ConvNext & 0.6327 & 0.4636 & 0.3602 & 0.2903 & 0.2487 & 0.4648 & 0.8272 \\
    ResNet & 0.6322 & 0.4621 & 0.3557 & 0.2835 & 0.2489 & 0.4649 & 0.8206 \\
    ResNext & 0.6246 & 0.4525 & 0.3463 & 0.2639 & 0.2294 & 0.4448 & 0.8117 \\
    DenseNet & 0.6298 & 0.4566 & 0.3504 & 0.2668 & 0.2207 & 0.4519 & 0.8167 \\
    ConvNext-ResNet & \textbf{0.6452} & \textbf{0.4767} & \textbf{0.3732} & \textbf{0.3087} & \textbf{0.2609} & \textbf{0.4827} & \textbf{0.8594} \\
    ConvNext-ResNext & 0.6414 & 0.4717 & 0.3676 & 0.2970 & 0.2551 & 0.4741 & 0.8429 \\
    ConvNext-DenseNet & 0.6392 & 0.4634 & 0.3699 & 0.3001 & 0.2588 & 0.4744 & 0.8392 \\
    ResNet-ResNext & 0.6427 & 0.4734 & 0.3684 & 0.2974 & 0.2558 & 0.4743 & 0.8431 \\
    ResNet-DenseNet & 0.6345 & 0.4578 & 0.3542 & 0.2741 & 0.2443 & 0.4694 & 0.8346 \\
    ResNext-DenseNet & 0.6282 & 0.4552 & 0.3519 & 0.2712 & 0.2311 & 0.4560 & 0.8211 \\
    \hline
\end{tabular}}}
\end{table}

The performance of various CNNs and their pairwise combinations is presented in~\Cref{SYDNEY_CNN,UCM_CNN,RSICD_CNN}. As shown in the results, the combination of features from two distinct CNN architectures consistently improves performance compared to using a single CNN alone. This fusion does not add significant computational cost, as CNNs are used solely for feature extraction and remain frozen during training. Since the encoder parameters are not updated, the training remains efficient while benefiting from the diverse spatial and semantic information captured by multiple CNNs. Among the combinations, ConvNext and ResNet yield the best results due to their complementary properties. ConvNext effectively captures global spatial patterns, while ResNet retains fine-grained details through residual connections. Previous studies \cite{das2025good,das2024unveiling} also highlight their superiority in RSIC tasks. Their fusion results in richer and more robust feature representations, making ConvNext and ResNet the chosen pair for all subsequent experiments in this study.
\subsection{Deep Decoding Strategy for Semantic Modeling}
\label{sec:stacking}
In this paper, we have used the GRU~\cite{khan2022deep} as a decoder. The GRU handles the flow of information through its gating mechanisms efficiently by retaining important temporal dependencies and being less computationally expensive. We have used a stacked GRU architecture in the decoder to improve the sequential modeling of the decoder. However, a trivial stacking (the output depends only on the last layer) may have the risk of discarding intermediate information and gradients~\cite{gu2018stack}. To deal with this, we examine four methods of combining the models as follows.
\subsubsection{Simple Stacking}
The traditional stacking approach, called Simple Stacking (SS), directly utilizes the output of the final decoder layer as the final output of the system. The corresponding equation of SS is given below.  
\begin{equation}  
    O = D_n  
\end{equation}
\subsubsection{Concatenation Stacking} 
In Concatenation Stacking (CS), the outputs of all decoder layers are concatenated to form the final output, thus preserving a richer set of feature representations. The corresponding equation of CS is given below.
\begin{equation}
    O = D_1|D_2|...|D_n
\end{equation}
\subsubsection{Global Weighted Stacking}  
In Global Weighted Stacking (GWS), a weighted average of the output of all decoder layers is calculated using a single global weight, ensuring a balanced contribution from each layer. The corresponding equation of GWS is given below.
\begin{equation}
    O = \sum_{i=1}^{n} \alpha_i{D_i}
\end{equation}
\subsubsection{Local Weighted Stacking}  
In Local Weighted Stacking (LWS), a weighted average of the decoder output is calculated using element-specific weights, allowing for finer control over the contribution of each feature to the final output. The corresponding equation of LWS is given below.
\begin{equation}
    O = \sum_{i=1}^{n} W_i \odot D_i 
\end{equation}
Here, $n$ denotes the number of decoders, and $D_i$ represents the output of the $i^{\text{th}}$ decoder, while $O$ is the final output of the stacked system. The symbol $|$ indicates the concatenation operation. The parameter $\alpha_i$ represents a scalar weight, while $W_i$ is a vector weight of the same size as $D_i$ for each $D_i$. Furthermore, $0 \leq \alpha_i \leq 1$ with constraint $\sum_{i=1}^{n} \alpha_i = 1$, and $0 \leq W_i^j \leq 1$ with $\sum_{i=1}^{n} W_i^j = 1$, where $W_i^j$ denotes the value of the characteristic $j^{\text{th}}$ of $W_i$. Both $\alpha = \{\alpha_i \mid i = 1, \dots, n\}$ and $W = \{W_i \mid i = 1, \dots, n\}$ are trainable parameters.
\begin{table}[!ht]
\centering
\caption{Comparison of Different Decoder Stacking Methods on the SYDNEY Dataset}
\label{sydney_stacked}
\resizebox{\linewidth}{!}{
    {\small
    \begin{tabular}{|ccccccccc|}
    \hline
    DC & Stack & BLEU-1 & BLEU-2 & BLEU-3 & BLEU-4 & METEOR & ROUGE-L & CIDEr \\
    \hline
    1 & NS & 0.7679 & 0.6814 & 0.6084 & 0.5458 & 0.3801 & 0.7263 & 2.4515 \\
    \hdashline
    \multirow{4}{*}{2}
    & SS & 0.7576 & 0.6825 & 0.6217 & 0.5672 & 0.4093 & 0.7097 & 2.4382 \\
    & CS & 0.7784 & 0.7012 & 0.6400 & 0.5853 & 0.4287 & 0.7319 & 2.5614 \\
    & GWS & 0.7856 & 0.7106 & 0.6482 & 0.5923 & 0.4195 & 0.7286 & 2.5691 \\
    & LWS & 0.8096 & 0.7397 & 0.6580 & 0.6114 & 0.4369 & 0.7461 & 2.6127 \\
    \hdashline
    \multirow{4}{*}{3}
    & SS & 0.7539 & 0.6773 & 0.6138 & 0.5587 & 0.4022 & 0.7042 & 2.3934 \\
    & CS & 0.7680 & 0.6901 & 0.6247 & 0.5654 & 0.4088 & 0.7139 & 2.4645 \\
    & GWS & 0.8049 & 0.7319 & 0.6696 & 0.6198 & 0.4348 & 0.7479 & 2.6658 \\
    & LWS & \textbf{0.8349} & \textbf{0.7589} & \textbf{0.6952} & \textbf{0.6369} & \textbf{0.4554} & \textbf{0.7767} & \textbf{2.9230} \\
    \hdashline
    \multirow{4}{*}{4}
    & SS & 0.7357 & 0.6473 & 0.5797 & 0.5186 & 0.3971 & 0.6974 & 2.2343 \\
    & CS & 0.7576 & 0.6758 & 0.6099 & 0.5520 & 0.3993 & 0.7017 & 2.3367 \\
    & GWS & 0.7953 & 0.7212 & 0.6640 & 0.6134 & 0.4266 & 0.7403 & 2.6238 \\
    & LWS & 0.8144 & 0.7423 & 0.6849 & 0.6338 & 0.4466 & 0.7587 & 2.7044 \\
    \hdashline
    \multirow{4}{*}{5}
    & SS & 0.7506 & 0.6618 & 0.5925 & 0.5331 & 0.4058 & 0.7016 & 2.2981 \\
    & CS & 0.7646 & 0.6769 & 0.6136 & 0.5577 & 0.4039 & 0.7136 & 2.4011 \\
    & GWS & 0.7998 & 0.7282 & 0.6668 & 0.6111 & 0.4288 & 0.7429 & 2.5669 \\
    & LWS & 0.8276 & 0.7537 & 0.6904 & 0.6306 & 0.4509 & 0.7706 & 2.7253 \\
    \hline
\end{tabular}}}
\end{table}
\begin{table}[!ht]
\centering
\caption{Comparison of Different Decoder Stacking Methods on the UCM Dataset}
\label{ucm_stacked}
\resizebox{\linewidth}{!}{
    \begin{tabular}{|ccccccccc|}
    \hline
    DC & Stack & BLEU-1 & BLEU-2 & BLEU-3 & BLEU-4 & METEOR & ROUGE-L & CIDEr \\
    \hline
    1 & NS & 0.8457 & 0.7919 & 0.7533 & 0.7166 & 0.4833 & 0.8163 & 3.4255 \\
    \hdashline
    \multirow{4}{*}{2}
    & SS & 0.8366 & 0.7764 & 0.7247 & 0.6771 & 0.4558 & 0.7904 & 3.4532 \\
    & CS & 0.8416 & 0.7860 & 0.7379 & 0.6910 & 0.4642 & 0.8038 & 3.5311 \\
    & GWS & 0.8572 & 0.8012 & 0.7510 & 0.7028 & 0.4745 & 0.8089 & 3.4716 \\
    & LWS & 0.8663 & 0.8157 & 0.7711 & 0.7276 & 0.4948 & 0.8307 & 3.5485 \\
    \hdashline
    \multirow{4}{*}{3}
    & SS & 0.8229 & 0.7606 & 0.7080 & 0.6590 & 0.4528 & 0.7835 & 3.4372 \\
    & CS & 0.8398 & 0.7848 & 0.7356 & 0.6881 & 0.4622 & 0.8019 & 3.5216 \\
    & GWS & 0.8678 & 0.8158 & 0.7706 & 0.7266 & 0.4889 & 0.8265 & 3.6196 \\
    & LWS & \textbf{0.8805} & \textbf{0.8305} & \textbf{0.7862} & \textbf{0.7511} & \textbf{0.5014} & \textbf{0.8497} & \textbf{3.6806} \\
    \hdashline
    \multirow{4}{*}{4}
    & SS & 0.8163 & 0.7583 & 0.7096 & 0.6643 & 0.4481 & 0.7842 & 3.4467 \\
    & CS & 0.8397 & 0.7844 & 0.7374 & 0.6897 & 0.4685 & 0.8072 & 3.4939 \\
    & GWS & 0.8612 & 0.8058 & 0.7581 & 0.7116 & 0.4769 & 0.8173 & 3.5104 \\
    & LWS & 0.8661 & 0.8090 & 0.7614 & 0.7170 & 0.4849 & 0.8215 & 3.5838 \\
    \hdashline
    \multirow{4}{*}{5}
    & SS & 0.8152 & 0.7551 & 0.7044 & 0.6558 & 0.4407 & 0.7742 & 3.3914 \\
    & CS & 0.8361 & 0.7818 & 0.7376 & 0.6915 & 0.4633 & 0.7958 & 3.4776 \\
    & GWS & 0.8468 & 0.7904 & 0.7416 & 0.6952 & 0.4806 & 0.8143 & 3.4959 \\
    & LWS & 0.8613 & 0.8054 & 0.7593 & 0.7157 & 0.4896 & 0.8147 & 3.6018 \\
    \hline
\end{tabular}}
\end{table}
\begin{table}[!ht]
\centering
\caption{Comparison of Different Decoder Stacking Methods on the RSICD Dataset}
\label{rsicd_stacked}
\resizebox{\linewidth}{!}{
    {\small
    \begin{tabular}{|ccccccccc|}
    \hline
    DC & Stack & BLEU-1 & BLEU-2 & BLEU-3 & BLEU-4 & METEOR & ROUGE-L & CIDEr \\
    \hline
    1 & NS & 0.6452 & 0.4767 & 0.3732 & 0.3087 & 0.2609 & 0.4827 & 0.8594 \\
    \hdashline
    \multirow{4}{*}{2}
    & SS & 0.6489 & 0.4786 & 0.3731 & 0.3012 & 0.2599 & 0.4778 & 0.8571 \\
    & CS & 0.6536 & 0.4871 & 0.3846 & 0.3146 & 0.2698 & 0.4888 & 0.8641 \\
    & GWS & 0.6586 & 0.4921 & 0.3881 & 0.3166 & 0.2712 & 0.4921 & 0.8822 \\
    & LWS & 0.6683 & 0.5007 & 0.3969 & 0.3265 & 0.2836 & 0.5027 & 0.8818 \\
    \hdashline
    \multirow{4}{*}{3}
    & SS & 0.6436 & 0.4759 & 0.3716 & 0.3003 & 0.2606 & 0.4782 & 0.8531 \\
    & CS & 0.6531 & 0.4869 & 0.3818 & 0.3116 & 0.2671 & 0.4894 & 0.8697 \\
    & GWS & 0.6614 & 0.5014 & 0.3984 & 0.3314 & 0.2911 & 0.5134 & 0.9196 \\
    & LWS & \textbf{0.6796} & \textbf{0.5124} & \textbf{0.4073} & \textbf{0.3365} & \textbf{0.2944} & \textbf{0.5248} & \textbf{0.9423} \\
    \hdashline
    \multirow{4}{*}{4}
    & SS & 0.6364 & 0.4687 & 0.3652 & 0.2956 & 0.2578 & 0.4727 & 0.8363 \\
    & CS & 0.6463 & 0.4797 & 0.3757 & 0.3043 & 0.2530 & 0.4698 & 0.8441 \\
    & GWS & 0.6511 & 0.4846 & 0.3806 & 0.3091 & 0.2637 & 0.4846 & 0.8747 \\
    & LWS & 0.6596 & 0.4924 & 0.3873 & 0.3165 & 0.2744 & 0.4938 & 0.8823 \\
    \hdashline
    \multirow{4}{*}{5}
    & SS & 0.6379 & 0.4683 & 0.3627 & 0.2906 & 0.2566 & 0.4710 & 0.8250 \\
    & CS & 0.6425 & 0.4734 & 0.3726 & 0.3014 & 0.2490 & 0.4636 & 0.8361 \\
    & GWS & 0.6502 & 0.4812 & 0.3745 & 0.2894 & 0.2511 & 0.4511 & 0.8426 \\
    & LWS & 0.6519 & 0.4843 & 0.3800 & 0.3089 & 0.2692 & 0.4886 & 0.8653 \\
    \hline
\end{tabular}}}
\end{table}

The impact of the depth of the decoder is evaluated in~\Cref{sydney_stacked,ucm_stacked,rsicd_stacked}, where the number of stacked GRUs (DC) is treated as a hyperparameter ranging from two to five. The number of decoders larger than five is excluded as they become computationally more expensive and yield lower returns in exchange.

Deeper stacking for SS and CS in the SYDNEY and UCM datasets(\Cref{sydney_stacked,ucm_stacked}) has inconsistent trends that lead to performance degradation when the number of layers exceeds two. SS performs particularly worse than the single decoder baseline (NS), possibly due to vanishing gradients and the small size of the training data~\cite{lu2017exploring}. In contrast, the RSICD dataset (\Cref{rsicd_stacked}) performs stably until the depth of three layers and declines beyond that, causing the performance insensitivity to the depth of the decoder in this dataset. From the results, we observe that among the stacking strategies, SS and CS are less effective. SS eliminates intermediate outputs, while CS introduces redundancy by treating all outputs equally. Weighted strategies offer a better balance: GWS uses global weights but lacks temporal flexibility, whereas LWS applies adaptive element-wise weights, enabling finer control over temporal dependencies. LWS consistently achieves the best results, with three stacked decoders yielding optimal performance across all datasets.
\section{Experimental Setup}
\label{sec:exp_setup}
We use a conventional encoder-decoder architecture. Each GRU on the decoder stack ($G$) has an output size of $256$. Linear layers $L1$, $L2$, and $L3$ have output sizes of $256$, $512$, and one more than the vocabulary size, respectively. GELU is used as the activation function for $L1$ and $L2$, while $L3$ uses \emph{SoftMax}. A dropout rate of $0.5$ is applied before $L1$, after the embedding layer ($E$), before the stack of linear layers ($SL$), and between consecutive GRUs to prevent overfitting. The embedding size is set to $256$.

Training is carried out with a batch size of $64$ using the Adam optimizer (learning rate $10^{-4}$) and categorical cross entropy as a loss function. An early stopping criterion with a patience of eight epochs is used, monitoring ROUGE-L on the validation set. For caption generation, we apply the comparison-based beam search \cite{das2024textgcn,hoxha2020new} with a beam width of five, using four similar images. The comparison score is calculated as the arithmetic mean of BLEU-2, METEOR, and ROUGE-L.

The experiments were carried out using TensorFlow within a Docker environment with GPU support, on a host machine equipped with a \emph{NVIDIA RTX A6000} GPU with $50GB$ memory and $132GB$ RAM. During manuscript preparation, ChatGPT and Grammarly were used exclusively for rephrasing and language refinement, without generating original technical content.
\subsection{Datasets Used}
\begin{itemize}
    \item\textbf{SYDNEY:} SYDNEY \cite{qu2016deep} dataset consists of $613$ images, with 497 allocated for training, $58$ for testing, and $58$ for validation. It is derived from the Sydney dataset \cite{zhang2014saliency} and has been carefully curated to include seven distinct categories.
    \item\textbf{UCM:} UCM \cite{qu2016deep} dataset consists of $2,100$ images, with $1,680$ allocated for training, $210$ for testing, and $21$0 for validation. It is derived from the \emph{UC Merced Land Use} dataset \cite{yang2010bag}, which features $21$ land use image classes, each containing $100$ images.
    \item\textbf{RSICD:} RSICD \cite{lu2017exploring} dataset includes a substantial collection of $10921$ images, with $8{,}034$ designated for training, $1{,}093$ for testing and $1{,}094$ for validation. Sourced from a variety of platforms such as Google Earth \cite{xia2017aid}, Baidu Map, MapABC, and Tianditu, this dataset encompasses $31$ distinct image classes.
\end{itemize}

In this study, we used the corrected versions of these datasets \cite{das2024textgcn}, which address common issues such as spelling mistakes, grammatical errors, and inconsistent dialect variations of English. We used the same train-validation-test split as defined in these datasets.
\subsection{Performance metrics}
\begin{table}[!ht]
    \centering
    \caption{Experimental Results of Several RSIC Methods on the SYDNEY Dataset}
    \label{SYDNEY_SEARCH}
    \resizebox{\linewidth}{!}{
    {\small
    \begin{tabular}{|cccccccc|}
    \hline
    METHOD & BLEU-1 & BLEU-2 & BLEU-3 & BLEU-4 & METEOR & ROUGE-L & CIDEr \\
    \hline
    R-BOW \cite{lu2017exploring} & 0.5310 & 0.4076 & 0.3319 & 0.2788 & 0.2490 & 0.4922 & 0.7019 \\
    L-FV \cite{lu2017exploring} & 0.6331 & 0.5332 & 0.4735 & 0.4303 & 0.2967 & 0.5794 & 1.4760 \\
    CSMLF \cite{wang2019semantic} & 0.4441 & 0.3369 & 0.2815 & 0.2408 & 0.1576 & 0.4018 & 0.9378 \\
    CSMLF-FT \cite{wang2019semantic} & 0.5998 & 0.4583 & 0.3869 & 0.3433 & 0.2475 & 0.5018 & 0.7555 \\
    SVM-DBOW \cite{hoxha2021novel} & 0.7787 & 0.6835 & 0.6023 & 0.5305 & 0.3797 & 0.6992 & 2.2722 \\
    SVM-DCONC \cite{hoxha2021novel} & 0.7547 & 0.6711 & 0.5970 & 0.5308 & 0.3643 & 0.6746 & 2.2222 \\
    TrTr-CMR \cite{wu2024trtr} & 0.8270 & 0.6994 & 0.6002 & 0.5199 & 0.3803 & 0.7220 & 2.2728 \\
    TextGCN \cite{das2024textgcn} & 0.7680 & 0.6892 & 0.6261 & 0.5786 & 0.4009 & 0.7314 & 2.8595 \\
    MsEdF & \textbf{0.8349} & \textbf{0.7589} & \textbf{0.6952} & \textbf{0.6369} & \textbf{0.4554} & \textbf{0.7767} & \textbf{2.9230} \\
    \hline
    \end{tabular}}}
\end{table}
To assess the performance of our model, we use widely used evaluation metrics in RSIC tasks, including BLEU \cite{papineni-etal-2002-bleu}, METEOR \cite{lavie-agarwal-2007-meteor}, ROUGE \cite{lin-2004-ROUGE}, and CIDEr \cite{vedantam2015cider}. 
\begin{itemize}
    \item\textbf{BLEU:} The Bilingual Evaluation Understudy (BLEU) metric \cite{papineni-etal-2002-bleu} assesses the quality of the generated text by measuring the n-gram overlap between a generated caption and its reference captions. It calculates the geometric mean of n-gram precision scores and applies a brevity penalty to discourage excessively short outputs. BLEU is widely used in tasks such as machine translation and image captioning. In this work, we evaluate the performance using BLEU-1 through BLEU-4.
    \item\textbf{METEOR:} The Metric for Evaluation of Translation with Explicit Ordering (METEOR) \cite{lavie-agarwal-2007-meteor} is an evaluation metric that accounts for stemming, synonymy and word order to assess the similarity between generated and reference captions. Unlike BLEU, which is based purely on precision, METEOR incorporates both precision and recall, with a stronger emphasis on accuracy.
    \item\textbf{ROUGE:} ROUGE \cite{lin-2004-ROUGE} stands for \emph{Recall-Oriented Understudy for Gisting Evaluation}. It is a recall-based metric that measures the overlap in n-grams between a generated caption and reference captions. In this work, we use ROUGE-L, which is based on the longest common subsequence (LCS).
    \item\textbf{CIDEr:} CIDEr \cite{vedantam2015cider} stands for \emph{Consensus-based Image Description Evaluation}. It is a metric used to evaluate the quality of generated captions for images by comparing them with a set of human-written reference captions. CIDEr measures how well a generated caption aligns with the collective understanding of the image across multiple human descriptions, aiming to capture the similarity between a generated caption and the general human perspective on the image content.
\end{itemize}
\section{Experimental Results}
\label{sec:results}
This section presents a comprehensive evaluation of the proposed model, including quantitative comparisons with baseline methods and ablation studies, followed by qualitative analysis through test image visualizations and subjective evaluation.
\subsection{Numerical Evaluation}
\begin{table}[!ht]
    \centering
    \caption{Experimental Results of Several RSIC Methods on the UCM Dataset}
    \label{UCM_SEARCH}
    \resizebox{\linewidth}{!}{
    {\small
    \begin{tabular}{|cccccccc|}
    \hline
    METHOD & BLEU-1 & BLEU-2 & BLEU-3 & BLEU-4 & METEOR & ROUGE-L & CIDEr \\
    \hline
    R-BOW \cite{lu2017exploring} & 0.4107 & 0.2249 & 0.1452 & 0.1095 & 0.1098 & 0.3439 & 0.3071 \\
    L-FV \cite{lu2017exploring} & 0.5897 & 0.4668 & 0.4080 & 0.3683 & 0.2698 & 0.5595 & 1.8438 \\
    CSMLF \cite{wang2019semantic} & 0.3874 & 0.2145 & 0.1253 & 0.0915 & 0.0954 & 0.3599 & 0.3703 \\
    CSMLF-FT \cite{wang2019semantic} & 0.3671 & 0.1485 & 0.0763 & 0.0505 & 0.0944 & 0.2986 & 0.1351 \\
    SVM-DBOW \cite{hoxha2021novel} & 0.7635 & 0.6664 & 0.5869 & 0.5195 & 0.3654 & 0.6801 & 2.7142 \\
    SVM-DCONC \cite{hoxha2021novel} & 0.7653 & 0.6947 & 0.6417 & 0.5942 & 0.3702 & 0.6877 & 2.9228 \\
    TrTr-CMR \cite{wu2024trtr} & 0.8156 & 0.7091 & 0.6220 & 0.5469 & 0.3978 & 0.7442 & 2.4742 \\
    TextGCN \cite{das2024textgcn} & 0.8461 & 0.7844 & 0.7386 & 0.6930 & 0.4868 & 0.8071 & 3.4077 \\
    MsEdF & \textbf{0.8805} & \textbf{0.8305} & \textbf{0.7862} & \textbf{0.7511} & \textbf{0.5014} & \textbf{0.8497} & \textbf{3.6806} \\
    \hline
    \end{tabular}}}
\end{table}
\begin{table}[!ht]
    \centering
    \caption{Experimental Results of Several RSIC Methods on the RSICD Dataset}
    \label{RSICD_SEARCH}
    \resizebox{\linewidth}{!}{
    {\small
    \begin{tabular}{|cccccccc|}
    \hline
    METHOD & BLEU-1 & BLEU-2 & BLEU-3 & BLEU-4 & METEOR & ROUGE-L & CIDEr \\
    \hline
    R-BOW \cite{lu2017exploring} & 0.4401 & 0.2383 & 0.1514 & 0.1041 & 0.1684 & 0.3605 & 0.4667 \\
    L-FV \cite{lu2017exploring} & 0.4342 & 0.2453 & 0.1634 & 0.1175 & 0.1712 & 0.3818 & 0.6531 \\
    CSMLF \cite{wang2019semantic} & 0.5759 & 0.3859 & 0.2832 & 0.2217 & 0.2128 & 0.4455 & 0.5297 \\
    CSMLF-FT \cite{wang2019semantic} & 0.5106 & 0.2911 & 0.1903 & 0.1352 & 0.1693 & 0.3789 & 0.3388 \\
    SVM-DBOW \cite{hoxha2021novel} & 0.6112 & 0.4277 & 0.3153 & 0.2411 & 0.2303 & 0.4588 & 0.6825 \\
    SVM-DCONC \cite{hoxha2021novel} & 0.5999 & 0.4347 & 0.3355 & 0.2689 & 0.2299 & 0.4577 & 0.6854 \\
    TrTr-CMR \cite{wu2024trtr} & 0.6201 & 0.3937 & 0.2671 & 0.1932 & 0.2399 & 0.4895 & 0.7518 \\
    TextGCN \cite{das2024textgcn} & 0.6513 & 0.4819 & 0.3747 & 0.3085 & 0.2752 & 0.4804 & 0.8266 \\
    MsEdF & \textbf{0.6796} & \textbf{0.5124} & \textbf{0.4073} & \textbf{0.3365} & \textbf{0.2944} & \textbf{0.5248} & \textbf{0.9423} \\
    \hline
    \end{tabular}}}
\end{table}

\Cref{SYDNEY_SEARCH,UCM_SEARCH,RSICD_SEARCH} compares various baseline methods. R-BOW and L-FV \cite{lu2017exploring} use handcrafted sentence representations with RNN and LSTM decoders, respectively. CSMLF and its fine-tuned variant \cite{wang2019semantic} extend this for multi-sentence captioning. SVM-DBOW and SVM-DCONC \cite{hoxha2021novel} apply SVMs with bag-of-words and concatenated sentence features. TrTr-CMR \cite{wu2024trtr} combines a Swin Transformer encoder with a cross-modal transformer decoder. TextGCN \cite{das2024textgcn} aligns the embeddings of the decoder with the frozen TextGCN output. Our model, MsEdF, combines dual CNN encoders with stacked GRUs and local weighted aggregation to enhance semantic representation. This method improves performance by effectively integrating the encoding and decoding strategies. The results demonstrate that the proposed MsEdF model achieves substantially superior performance compared to the baseline methods.
\subsection{Ablation Studies}
\begin{table}[!ht]
    \centering
    \caption{Performance Comparison of Model Variants Through Ablation Studies on the SYDNEY Dataset}
    \label{SYDNEY_ABLATION}
    \resizebox{\linewidth}{!}{
    {\small
    \begin{tabular}{|cccccccc|}
    \hline
    METHOD & BLEU-1 & BLEU-2 & BLEU-3 & BLEU-4 & METEOR & ROUGE-L & CIDEr \\
    \hline
    C-NS & 0.7531 & 0.6645 & 0.5982 & 0.5376 & 0.3750 & 0.6951 & 2.3954 \\
    R-NS & 0.7363 & 0.6502 & 0.5785 & 0.5138 & 0.3621 & 0.6809 & 2.2526 \\
    FE-NS & 0.7679 & 0.6814 & 0.6084 & 0.5458 & 0.3801 & 0.7263 & 2.4515 \\
    C-LWS & 0.8141 & 0.7321 & 0.6610 & 0.6054 & 0.4369 & 0.7584 & 2.7472 \\
    R-LWS & 0.8056 & 0.7211 & 0.6527 & 0.5924 & 0.4216 & 0.7364 & 2.5743 \\
    FE-LWS & \textbf{0.8349} & \textbf{0.7589} & \textbf{0.6952} & \textbf{0.6369} & \textbf{0.4554} & \textbf{0.7767} & \textbf{2.9230} \\
    \hline
    \end{tabular}}}
\end{table}
\begin{table}[!ht]
    \centering
    \caption{Performance Comparison of Model Variants Through Ablation Studies on the UCM dataset}
    \label{UCM_ABLATION}
    \resizebox{\linewidth}{!}{
    {\small
    \begin{tabular}{|cccccccc|}
    \hline
    METHOD & BLEU-1 & BLEU-2 & BLEU-3 & BLEU-4 & METEOR & ROUGE-L & CIDEr \\
    \hline
    C-NS & 0.8369 & 0.7804 & 0.7312 & 0.6853 & 0.4689 & 0.8037 & 3.3990 \\
    R-NS & 0.8297 & 0.7705 & 0.7178 & 0.6683 & 0.4506 & 0.7903 & 3.3329 \\
    FE-NS & 0.8457 & 0.7919 & 0.7533 & 0.7166 & 0.4833 & 0.8163 & 3.4255 \\
    C-LWS & 0.8655 & 0.8121 & 0.7644 & 0.7192 & \textbf{0.5086} & 0.8367 & 3.4958 \\
    R-LWS & 0.8614 & 0.8034 & 0.7526 & 0.7049 & 0.4910 & 0.8194 & 3.4621 \\
    FE-LWS & \textbf{0.8805} & \textbf{0.8305} & \textbf{0.7862} & \textbf{0.7511} & 0.5014 & \textbf{0.8497} & \textbf{3.6806} \\
    \hline
    \end{tabular}}}
\end{table}
\begin{table}[!ht]
    \centering
    \caption{Performance Comparison of Model Variants Through Ablation Studies on the RSICD dataset}
    \label{RSICD_ABLATION}
    \resizebox{\linewidth}{!}{
    {\small
    \begin{tabular}{|cccccccc|}
    \hline
    METHOD & BLEU-1 & BLEU-2 & BLEU-3 & BLEU-4 & METEOR & ROUGE-L & CIDEr \\
    \hline
    C-NS & 0.6327 & 0.4636 & 0.3602 & 0.2903 & 0.2487 & 0.4648 & 0.8272 \\
    R-NS & 0.6322 & 0.4621 & 0.3557 & 0.2835 & 0.2489 & 0.4649 & 0.8206 \\
    FE-NS & 0.6452 & 0.4767 & 0.3732 & 0.3087 & 0.2609 & 0.4827 & 0.8594 \\
    C-LWS & 0.6671 & 0.5008 & 0.3996 & 0.3302 & 0.2843 & 0.5066 & 0.8954 \\
    R-LWS & 0.6645 & 0.4981 & 0.3963 & 0.3257 & 0.2849 & 0.5010 & 0.8684 \\
    FE-LWS & \textbf{0.6796} & \textbf{0.5124} & \textbf{0.4073} & \textbf{0.3365} & \textbf{0.2944} & \textbf{0.5248} & \textbf{0.9423} \\
    \hline
    \end{tabular}}}
\end{table}
To evaluate the contribution of individual components, we compare four encoder-decoder configurations: single encoder with single decoder, fusion-based encoder with single decoder, single encoder with LWS decoder and the proposed fusion-based encoder with LWS decoder. The results presented in~\Cref{SYDNEY_ABLATION,UCM_ABLATION,RSICD_ABLATION} follow the naming convention: \emph{C} $\rightarrow$ ConvNext, \emph{R} $\rightarrow$ ResNet and \emph{FE} $\rightarrow$ Fusion of \emph{C} and \emph{R}. For decoder configurations, \emph{NS} $\rightarrow$ a single decoder (non-stacked), while \emph{LWS} $\rightarrow$ Local Weighted Stacking using three GRUs (see~\Cref{sec:stacking} for details), \emph{FE-LWS} $\rightarrow$ the proposed method \emph{MsEdF}. The results indicate that both encoder fusion and decoder stacking improve performance, with the MsEdF configuration achieving the highest scores across all datasets. In particular, LWS offers greater performance gains than encoder fusion alone, highlighting its effectiveness even when used with a single CNN.
\subsection{Visual Interpretation}
\begin{figure}[!ht]
  \centering
  \subfloat[\label{example1}]{\includegraphics[width=0.33\linewidth,height=150px]{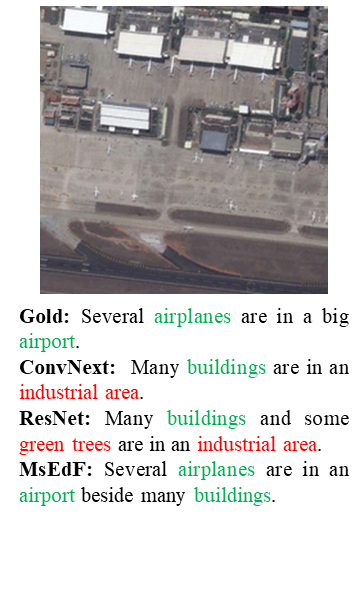}}
  \subfloat[\label{example2}]{\includegraphics[width=0.33\linewidth,height=150px]{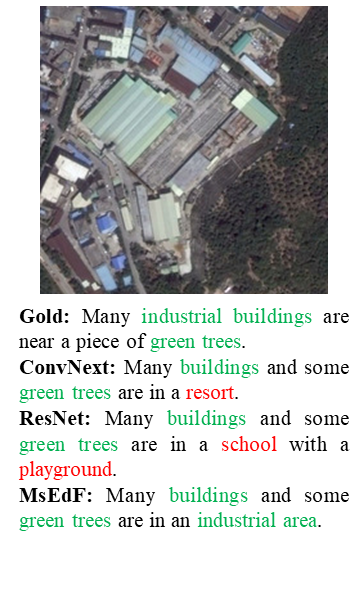}}
  \subfloat[\label{example3}]{\includegraphics[width=0.33\linewidth,height=150px]{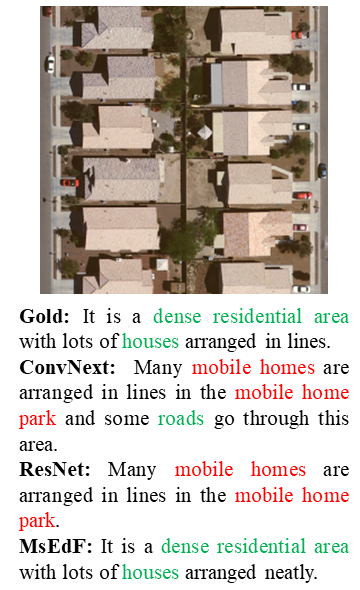}}
  
  \subfloat[\label{example7}]{\includegraphics[width=0.33\linewidth,height=150px]{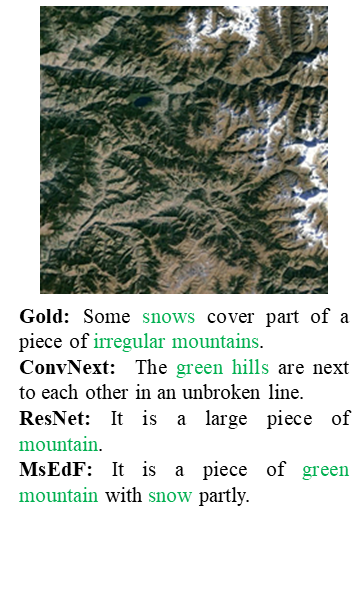}}
  \subfloat[\label{example8}]{\includegraphics[width=0.33\linewidth,height=150px]{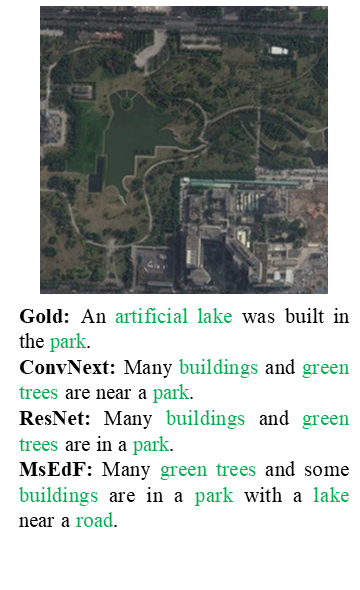}}
  \subfloat[\label{example9}]{\includegraphics[width=0.33\linewidth,height=150px]{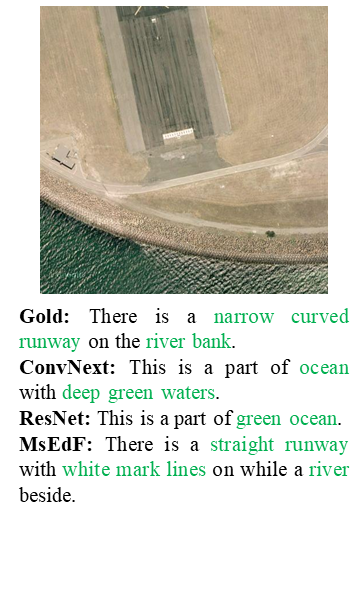}}
   \caption{Examples of RS Image Captioning by Different Methods}
   \label{fig_visual}
\end{figure}
We present qualitative results from test images in~\Cref{fig_visual}. Here we have compared the proposed model (MsEdF) with ground-truth captions (Gold) and conventional encoder-decoder models (ConvNext and ResNet). These examples highlight the cases where the proposed approach produces captions that are semantically rich. 
Several examples (\Cref{example1,example2,example3}) show cases where the baseline model produces incorrect or misleading captions, whereas the proposed model generates accurate and contextually relevant descriptions.
In~\Cref{example1}, ConvNext and ResNet misclassify the image as \emph{industrial area}.
In~\Cref{example2}, ConvNext misclassifies the image as \emph{resort} and ResNet as \emph{school}.
In~\Cref{example3}, ConvNext and ResNet misclassify the image as \emph{mobile home park}.
Other example (\Cref{example7,example8,example9}) highlights that, although both models produce generally correct captions, the proposed approach incorporates extra visual or semantic details that enhance the overall quality of the caption.
In~\Cref{example7}, while all models identify \emph{mountain}, the proposed model further recognizes \emph{snow}.
In~\Cref{example8}, while all models identify \emph{park}, the proposed model further recognizes \emph{lake} and \emph{road}, enhancing the semantic richness of the caption.
In~\Cref{example9}, while all models identify \emph{ocean} or \emph{river} (Visually, it is challenging to distinguish whether it is an \emph{ocean} or an \emph{river}; therefore, both interpretations are accepted), the proposed model also recognizes \emph{runway} with \emph{marking lines}.
\subsection{Subjective Evaluation}
\label{sec:subjective}
Unlike classification tasks, image captioning allows multiple valid descriptions of the same image. Therefore, evaluating the generated captions solely based on the five reference captions provided in each of the three datasets is insufficient \cite{das2024unveiling,lu2017exploring}. To address this, we conducted a qualitative evaluation using a human annotator.\footnote{A domain expert with several years of experience in RSIC.} This human-centric assessment provides a deeper understanding of the quality of captions. In our study, we used three evaluation labels: Related, where the caption accurately describes the input image; Partially Related, where the caption has significant flaws but still relates to the image; and Unrelated, where the caption does not describe the image meaningfully.
\begin{table*}[!ht]
\centering
\caption{Subjective Evaluation of Different Encoder-decoder models on Three Datasets (in \%)}
\label{subjective}
\resizebox{\linewidth}{!}{%
\begin{tabular}{|c|c|c|c|c|c|c|c|c|c|}
\hline
\multirow{2}{*}{Method} & \multicolumn{3}{c|}{SYDNEY} & \multicolumn{3}{c|}{UCM} & \multicolumn{3}{c|}{RSICD} \\ 
\cline{2-10}
& Related & Partially Related & Unrelated & Related & Partially Related & Unrelated & Related & Partially Related & Unrelated \\
\hline
ConvNext & 89.66 & 1.72 & 8.62 & 89.04 & 4.29 & 6.67 & 82.25 & 5.31 & 12.44 \\
ResNet & 87.93 & 3.45 & 8.62 & 88.09 & 5.24 & 6.67 & 81.43 & 5.03 & 13.54 \\
MsEdF & 94.83 & 3.45 & 1.72 & 92.86 & 5.24 & 1.90 & 83.72 & 6.95 & 9.33 \\
\hline
\end{tabular}%
}
\end{table*}

The results of the subjective evaluation of different models are presented in~\Cref{subjective}. Here we compare our model (MsEdF) with conventional encoder and decoder models using ConvNext and ResNet encoders. The values indicate the percentage of test images labeled Related, Partially Related, or Unrelated based on human judgment. The results show that our model outperforms the baseline models in terms of perceptual accuracy.
\subsection{Error Analysis}
\label{error analysis}
During visual interpretation of different models, several common issues were identified in the generated captions. The most common problem is misclassification \cite{lu2017exploring}, where similar objects are confused. For example: \emph{industrial buildings} as \emph{resort} or \emph{school} (\Cref{example2}), \emph{dense residential area} as \emph{mobile home parks} (\Cref{example3}), \emph{river} as \emph{pond} and \emph{farmland} as \emph{bareland}. A more subtle error involves contextual confusion, where the presence of an object affects the overall interpretation of the image. For example, \emph{buildings} within the \emph{airport} image causes a misclassification of the image as \emph{industrial area} (\Cref{example1}). In some images, the main object is missing. An example is the absence of a \emph{runway} (\Cref{example9}). Minor object omissions also occur, such as missing \emph{houses} in \emph{meadow}, \emph{green trees} in \emph{bridge} image, \emph{snow} in \emph{mountains} (\Cref{example7})), and \emph{lake} in \emph{park} (\Cref{example8}). Additional issues include attribute misclassification (such as count, color), falsely adding non-existent objects, and incomplete sentence generation.
\section{Conclusion}
\label{sec:conclusion}
Remote sensing images are challenging to interpret in the form of captions due to complex spatial layouts and ambiguous semantic objects. In this work, we propose a multi-stream encoder-decoder framework for RSIC. Two structurally diverse CNN encoders capture the varied visual characteristics present in remote sensing images. On the decoder side, a stacked GRU-based decoder is used to capture hierarchical linguistic patterns and improve the modeling of long-term dependencies. Furthermore, we show that locally weighted layer aggregation enhances semantic modeling on the decoder and enables more context-aware captioning. Experimental results demonstrate that the proposed framework consistently improves caption quality over single-stream architectures by promoting feature diversity through encoder fusion and enhancing semantic modeling via decoder-side aggregation. However, our system may fail for images that are substantially different from those encountered during training. In future work, our aim is to improve our model for unseen images through retrieval-guided decoding or domain-adaptive training.
\bibliographystyle{splncs04} 
\bibliography{MyBib}

\begin{thebibliography}{10}
\providecommand{\url}[1]{\texttt{#1}}
\providecommand{\urlprefix}{URL }
\providecommand{\doi}[1]{https://doi.org/#1}

\bibitem{alghamdi2024predicting}
Alghamdi, M.A., Abdullah, S., Ragab, M.: Predicting energy consumption using stacked lstm snapshot ensemble. Big Data Mining and Analytics  \textbf{7}(2),  247--270 (2024)

\bibitem{das2025good}
Das, S., Gupta, S., Kumar, K., Sharma, R.: Good representation, better explanation: Role of convolutional neural networks in transformer-based remote sensing image captioning. arXiv preprint arXiv:2502.16095  (2025)

\bibitem{das2024unveiling}
Das, S., Khandelwal, A., Sharma, R.: Unveiling the power of convolutional neural networks: A comprehensive study on remote sensing image captioning and encoder selection. In: 2024 International Joint Conference on Neural Networks (IJCNN). pp.~1--8. IEEE (2024)

\bibitem{das2024textgcn}
Das, S., Sharma, R.: A textgcn-based decoding approach for improving remote sensing image captioning. IEEE Geoscience and Remote Sensing Letters  (2024)

\bibitem{gu2018stack}
Gu, J., Cai, J., Wang, G., Chen, T.: Stack-captioning: Coarse-to-fine learning for image captioning. In: Proceedings of the AAAI conference on artificial intelligence. vol.~32 (2018)

\bibitem{he2016deep}
He, K., Zhang, X., Ren, S., Sun, J.: Deep residual learning for image recognition. In: Proceedings of the IEEE conference on computer vision and pattern recognition. pp. 770--778 (2016)

\bibitem{hoxha2021novel}
Hoxha, G., Melgani, F.: A novel svm-based decoder for remote sensing image captioning. IEEE Transactions on Geoscience and Remote Sensing  \textbf{60},  1--14 (2021)

\bibitem{hoxha2020new}
Hoxha, G., Melgani, F., Slaghenauffi, J.: A new cnn-rnn framework for remote sensing image captioning. In: 2020 Mediterranean and Middle-East Geoscience and Remote Sensing Symposium (M2GARSS). pp.~1--4. IEEE (2020)

\bibitem{huang2017densely}
Huang, G., Liu, Z., Van Der~Maaten, L., Weinberger, K.Q.: Densely connected convolutional networks. In: Proceedings of the IEEE conference on computer vision and pattern recognition. pp. 4700--4708 (2017)

\bibitem{khan2022deep}
Khan, R., Islam, M.S., Kanwal, K., Iqbal, M., Hossain, M.I., Ye, Z.: A deep neural framework for image caption generation using gru-based attention mechanism. arXiv preprint arXiv:2203.01594  (2022)

\bibitem{lavie-agarwal-2007-meteor}
Lavie, A., Agarwal, A.: {METEOR}: An automatic metric for {MT} evaluation with high levels of correlation with human judgments. In: Proceedings of the Second Workshop on Statistical Machine Translation. pp. 228--231. Association for Computational Linguistics, Prague, Czech Republic (Jun 2007), \url{https://aclanthology.org/W07-0734}

\bibitem{li2025feature}
Li, Y., Tao, C., Liu, M., Zhang, X., Wang, G., Zhang, T., Zhao, D., Wang, D.: Feature refinement and rethinking attention for remote sensing image captioning. Scientific Reports  \textbf{15}(1), ~8742 (2025)

\bibitem{lin-2004-ROUGE}
Lin, C.Y.: {ROUGE}: A package for automatic evaluation of summaries. In: Text Summarization Branches Out. pp. 74--81. Association for Computational Linguistics, Barcelona, Spain (Jul 2004), \url{https://aclanthology.org/W04-1013}

\bibitem{liu2025semantic}
Liu, M., Liu, J., Zhang, X.: Semantic-spatial feature fusion with dynamic graph refinement for remote sensing image captioning. arXiv preprint arXiv:2503.23453  (2025)

\bibitem{liu2022convnet}
Liu, Z., Mao, H., Wu, C.Y., Feichtenhofer, C., Darrell, T., Xie, S.: A convnet for the 2020s. In: Proceedings of the IEEE/CVF conference on computer vision and pattern recognition. pp. 11976--11986 (2022)

\bibitem{lu2017exploring}
Lu, X., Wang, B., Zheng, X., Li, X.: Exploring models and data for remote sensing image caption generation. IEEE Transactions on Geoscience and Remote Sensing  \textbf{56}(4),  2183--2195 (2017)

\bibitem{papineni-etal-2002-bleu}
Papineni, K., Roukos, S., Ward, T., Zhu, W.J.: {B}leu: a method for automatic evaluation of machine translation. In: Proceedings of the 40th Annual Meeting of the Association for Computational Linguistics. pp. 311--318. Association for Computational Linguistics, Philadelphia, Pennsylvania, USA (Jul 2002). \doi{10.3115/1073083.1073135}, \url{https://aclanthology.org/P02-1040}

\bibitem{qu2016deep}
Qu, B., Li, X., Tao, D., Lu, X.: Deep semantic understanding of high resolution remote sensing image. In: 2016 International conference on computer, information and telecommunication systems (Cits). pp.~1--5. IEEE (2016)

\bibitem{sattar2024remote}
Sattar, A., Assam, M., Alahmadi, T.J., Bhatti, U.A., Tang, H., Aamir, M.: Remote sensing based advance image captioning improved feature attention. In: 2024 7th International Conference on Pattern Recognition and Artificial Intelligence (PRAI). pp. 97--105. IEEE (2024)

\bibitem{vedantam2015cider}
Vedantam, R., Lawrence~Zitnick, C., Parikh, D.: Cider: Consensus-based image description evaluation. In: Proceedings of the IEEE conference on computer vision and pattern recognition. pp. 4566--4575 (2015)

\bibitem{wang2019semantic}
Wang, B., Lu, X., Zheng, X., Li, X.: Semantic descriptions of high-resolution remote sensing images. IEEE Geoscience and Remote Sensing Letters  \textbf{16}(8),  1274--1278 (2019)

\bibitem{wang2024remote}
Wang, J., Wang, B., Xi, J., Bai, X., Ersoy, O.K., Cong, M., Gao, S., Zhao, Z.: Remote sensing image captioning with sequential attention and flexible word correlation. IEEE Geoscience and Remote Sensing Letters  (2024)

\bibitem{wu2024trtr}
Wu, Y., Li, L., Jiao, L., Liu, F., Liu, X., Yang, S.: Trtr-cmr: Cross-modal reasoning dual transformer for remote sensing image captioning. IEEE Transactions on Geoscience and Remote Sensing  (2024)

\bibitem{wu1509fusing}
Wu, Z., Jiang, Y., Wang, X., Ye, H., Xue, X., Wang, J.: Fusing multi-stream deep networks for video classification. arxiv 2015. arXiv preprint arXiv:1509.06086  (2015)

\bibitem{xia2017aid}
Xia, G.S., Hu, J., Hu, F., Shi, B., Bai, X., Zhong, Y., Zhang, L., Lu, X.: Aid: A benchmark data set for performance evaluation of aerial scene classification. IEEE Transactions on Geoscience and Remote Sensing  \textbf{55}(7),  3965--3981 (2017)

\bibitem{xie2017aggregated}
Xie, S., Girshick, R., Doll{\'a}r, P., Tu, Z., He, K.: Aggregated residual transformations for deep neural networks. In: Proceedings of the IEEE conference on computer vision and pattern recognition. pp. 1492--1500 (2017)

\bibitem{xu2015show}
Xu, K.: Show, attend and tell: Neural image caption generation with visual attention. arXiv preprint arXiv:1502.03044  (2015)

\bibitem{yang2010bag}
Yang, Y., Newsam, S.: Bag-of-visual-words and spatial extensions for land-use classification. In: Proceedings of the 18th SIGSPATIAL international conference on advances in geographic information systems. pp. 270--279 (2010)

\bibitem{you2016image}
You, Q., Jin, H., Wang, Z., Fang, C., Luo, J.: Image captioning with semantic attention. In: Proceedings of the IEEE conference on computer vision and pattern recognition. pp. 4651--4659 (2016)

\bibitem{zhang2014saliency}
Zhang, F., Du, B., Zhang, L.: Saliency-guided unsupervised feature learning for scene classification. IEEE Transactions on Geoscience and Remote Sensing  \textbf{53}(4),  2175--2184 (2014)

\end{thebibliography}
\end{document}